\documentclass[sigconf,nonacm,balance=false]{acmart}
\setcopyright{acmlicensed}
\acmYear{2027}
\copyrightyear{2027}
\acmConference[WWW '27]{The ACM Web Conference 2027}{May 10--14, 2027}{Dublin, Ireland}
\acmDOI{}
\acmISBN{}
\usepackage{tabularx}
\usepackage{multirow}
\makeatletter
\renewcommand{\@parfont}{\normalfont\bfseries}
\renewcommand{\@subsubsecfont}{\sffamily\bfseries}
\makeatother
\graphicspath{{./}}
\begin{document}
\title{Benchmarking Task Takeover Across Agent Handoffs}
\settopmatter{authorsperrow=3}

\author{Wenyuan Li}
\affiliation{\institution{Hokkaido University}
\city{Sapporo}\country{Japan}}

\author{Guang Li}
\authornote{Correspondence to Guang Li
\texttt{<guang@lmd.ist.hokudai.ac.jp>}.}
\affiliation{\institution{Hokkaido University}
\city{Sapporo}\country{Japan}}

\author{Dongyuan Li}
\affiliation{\institution{The University of Tokyo}
\city{Tokyo}\country{Japan}}

\author{Keisuke Maeda}
\affiliation{\institution{Hokkaido University}
\city{Sapporo}\country{Japan}}

\author{Takahiro Ogawa}
\affiliation{\institution{Hokkaido University}
\city{Sapporo}\country{Japan}}

\author{Miki Haseyama}
\affiliation{\institution{Hokkaido University}
\city{Sapporo}\country{Japan}}
\begin{abstract}
As agents collaborate on longer tasks, responsibility may pass between models before the work is complete. The receiving model must build on prior decisions, observations, and actions while addressing the remaining requirements. Evaluating this transition requires connecting the information preserved at handoff to how the receiver continues the work, a relationship that task-success and memory-retention scores alone do not capture. We introduce the Handoff State Continuity Benchmark (HSC-Bench), comprising 1,039 tasks that evaluate two complementary aspects of task takeover: recovering relevant repository evidence and continuing from established conversational, tool, and file-work state. Comparisons across six receiving models show that structured handoff state improves repository evidence recovery for every model and continuation performance for most models. The strongest receivers differ between the two tracks, showing that evidence-recovery rankings alone do not characterize performance when continuing from inherited state. By combining these comparisons with checks of subsequent responses and actions, HSC-Bench evaluates both access to prior evidence and adherence to established commitments and completed work. This provides a basis for selecting receiving models and assessing handoff representations according to the work that must be carried forward.
\end{abstract}
\ccsdesc[500]{Computing methodologies~Artificial intelligence}
\keywords{Agent handoff, task takeover, state continuity, agent evaluation}

\maketitle

\noindent\textbf{Relevance Statement.}
This work addresses the Semantics and Knowledge track by studying
knowledge continuity across LLM agent handoffs.
HSC-Bench evaluates how structured task state supports evidence
recovery and continuation from prior decisions and actions.
These capabilities are essential for reliable Web-based agentic workflows.

\section{Introduction}

\begin{figure}[!t]
\centering
\includegraphics[width=\linewidth]{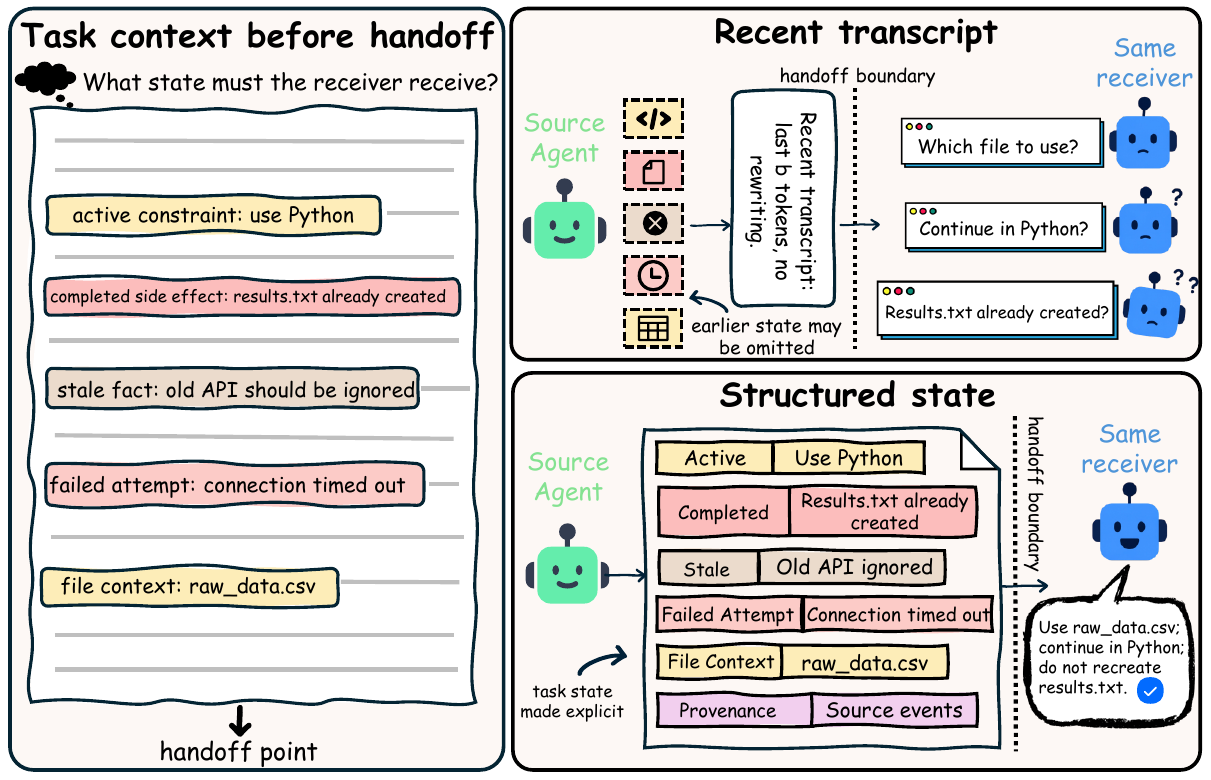}
\caption{Task takeover from a shared history.
A recent transcript may omit earlier decisions and completed
work, leaving the receiver uncertain about the next step.
Structured state preserves this information explicitly:
in the illustrated example, the receiver identifies the
required input, follows the established constraint, and
recognizes that an output has already been created.
The comparison holds the receiving model and visible token
count fixed.}
\Description{The motivating illustration shows a shared task history
and two handoff messages: a recent-history suffix and explicitly organized
task state, including a completed file creation that should not be repeated.}
\label{fig:overview}
\end{figure}

Web agents gather information and perform actions~\cite{shi2025autotools,cai2025puma,li2026deepagent}
across workflows that may involve multiple models~\cite{wu2024autogen,hong2024metagpt}.
When responsibility shifts before a task is finished, the receiving model
inherits a partially completed process: earlier interactions have established
decisions, accumulated evidence, and changed the environment.
These developments determine what remains to be done and which next steps
are appropriate. For example, a confirmed user choice should guide subsequent
work, while a completed service operation may no longer need to be performed.
We study \textbf{task takeover}: how a receiving model continues
another agent's unfinished work from the progress already made,
as illustrated in Figure~\ref{fig:overview}.

Evaluating task takeover requires connecting this inherited progress to the
receiver's subsequent behavior. Agent benchmarks assess task completion
and policy compliance~\cite{yao2025taubench,song2026colorbench}, while
interactive-memory benchmarks examine the retrieval and use of accumulated
or revised information~\cite{wu2025longmemeval}. These perspectives provide
relevant measures, but neither a final outcome nor successful recall alone
establishes whether a receiver has continued appropriately from the handoff
point. The evaluation must also account for what was already settled,
what remained unresolved, and how those conditions shaped the receiver's
response. This calls for making the state at handoff explicit and assessing
both the recovery of relevant evidence and the consistency of subsequent
responses and actions with that state.

We operationalize these requirements through two complementary evaluation
targets. \textbf{Evidence recovery} measures whether the receiver can identify
task-relevant information from the inherited context, using annotated
repository files and code spans as references~\cite{li2026contextbench}.
\textbf{State-dependent continuation} measures whether subsequent responses
and actions satisfy the requirements of the inherited task state, including
maintaining valid commitments, incorporating updated facts, and accounting
for completed operations. The two targets provide separate measures of
what a receiver recovers from prior work and how it continues from that work.

We instantiate these evaluation targets in HSC-Bench, a benchmark of
1,039 tasks. Its repository track pairs recorded coding-agent trajectories
with ContextBench annotations~\cite{li2026contextbench}, yielding 300 tasks
across 38 repositories. Its continuation track comprises 739 synthetic
tasks covering conversational state, tool use, and reported file-work
state, with explicit targets for evaluating subsequent responses and actions.

We evaluate six receiving models, including GPT-5.4 and Kimi K2,
using shared source histories and fixed handoff points, so that each
model takes over the same prior work. To examine the effect of inherited
context, we pair recent transcripts with structured state at the same
visible token count, holding the receiver and task request fixed.
This design supports comparisons across receiving models and within-model
estimates of how their performance changes with the context provided
at handoff.

The results show that evidence recovery and continuation reveal
complementary strengths of receiving models. Structured state improves
repository recovery across all evaluated models and continuation
performance for most, but the leading models differ between the two
tracks. A receiver's strength in locating relevant evidence therefore
provides only part of the basis for choosing it to continue unfinished
work. These findings motivate assessing handoff representations together
with the models that use them, judging their value by both the evidence
recovered and the subsequent responses and actions required by the
inherited task state.

To summarize, our contributions are as follows.
\begin{itemize}
    \item We introduce HSC-Bench, comprising 1,039 tasks that combine
    recorded repository work with synthetic continuation scenarios
    covering conversational state, tool use, and reported file-work state.

    \item We develop an evaluation protocol that scores recovered
    evidence against annotated code locations and checks whether
    subsequent responses and actions preserve valid commitments,
    incorporate updated facts, and respect completed operations.

    \item We show that structured state improves repository evidence
    recovery across all six receiving models and continuation performance
    for most, while the leading models differ between the two evaluation
    tracks, highlighting their complementary roles in assessing how
    receivers recover relevant evidence and continue from inherited
    task state.
\end{itemize}

\section{Related Work}

\subsection{Agent Evaluation}

Agent benchmarks evaluate whether models can complete tasks and satisfy
the requirements of their environments~\cite{liu2024agentbench,zhou2024webarena,drouin2024workarena}. SWE-bench examines issue
resolution in real repositories~\cite{jimenez2024swe}, while the Berkeley
Function Calling Leaderboard evaluates function selection and
execution~\cite{BFCL}. In stateful environments, success also depends
on how actions affect subsequent steps. ToolSandbox evaluates
intermediate and final milestones~\cite{lu2025toolsandbox}, and
$\tau$-bench checks the resulting database state together with
required user-facing information~\cite{yao2025taubench}. These benchmarks
connect agent behavior to observable outcomes and constraints.

More fine-grained evaluation examines which requirements an agent
satisfies within a task. This includes distinguishing task completion
from subtask progress and atomic capabilities in
ColorBench~\cite{song2026colorbench}, assessing scholarly search and
reading operations in PaperAsk~\cite{wu2026paperask}, and checking
retrieved product evidence against individual preferences in
AgenticShop~\cite{kim2026agenticshop}. We build on this emphasis on
specific requirements to evaluate a receiving model's continuation
of prior work. At a handoff, some requirements have already been
satisfied, while earlier decisions and actions constrain how the
remaining work should proceed. HSC-Bench evaluates the receiver
relative to this inherited progress, examining whether its subsequent
responses and actions satisfy the requirements that apply at the handoff.

\subsection{Interactive Memory}

Interactive memory enables agents to draw on information accumulated
across earlier interactions. Memory tiers, retrieval guided by global
memory, and history organization for personalized generation offer
different ways to make this information
accessible~\cite{packer2023memgpt,qian2025memorag,yu2026memweaver,zhong2024memorybank,maharana2024evaluating,wang2024agent}.
Evaluating memory also requires examining how retrieved information
is used~\cite{pan2025secom}. LongMemEval separates retrieval from reading and tests
reasoning across sessions, temporal changes, and knowledge
updates~\cite{wu2025longmemeval}. Related long-context studies show
that performance depends on evidence position and input length,
including when relevant information is successfully
retrieved~\cite{liu2024lost,du2025contextlength}. Together, these
studies motivate evaluating both access to prior information and
its contribution to subsequent responses.

Under a limited context budget, this distinction also shapes how
memory is compressed. Prompt compression selects which information
remains available~\cite{jiang2023llmlingua,jiang2024longllmlingua},
while failure-guided compression and continuation-based summary
validation assess how the retained information supports later
work~\cite{kang2025acon,chen2026slipstream}. Task takeover adds
a specific requirement to the use of prior information: earlier
decisions and completed actions establish what the receiving
model should do next. HSC-Bench examines whether the context
passed between agents supports continuation consistent with
these commitments and prior progress, connecting memory
evaluation to the requirements of an unfinished task.

\subsection{Evidence Recovery and Agent Handoffs}

Evaluating task takeover requires references for both the evidence
relevant to a task and the progress made before the handoff.
In repository work, annotated files and code spans provide a basis
for measuring evidence recovery~\cite{li2026contextbench}, while
budgeted context restoration evaluates retrieval with a shared
synthesis model~\cite{sheverev2026restoration}. Execution traces
complement these references by recording how the task reached
its current state; reconstructing state transitions can help
locate failures within that process~\cite{li2026codetracer}.
Our repository track uses ContextBench annotations to evaluate
the evidence recovered from recorded work at a fixed handoff point.

Studies of agent handoffs examine how prior work affects a model
that resumes an unfinished task. Dialogue and coding settings
reveal changes in continuation performance, rediscovery costs,
and issue resolution after a
handoff~\cite{khraishi2026evaluating,kc2026handoffdebt}.
Checkpoint-based evaluation further distinguishes reaching
a useful task state from successfully solving the task from
that state~\cite{liu2026reachsolve}. Our focus is on comparing
receiving models under shared handoff conditions. HSC-Bench
holds the source history and handoff point fixed across receivers
and evaluates two complementary targets: recovering annotated
repository evidence and satisfying explicit requirements when
continuing from inherited task state. This provides a common
basis for examining how receiver performance differs between
identifying relevant prior information and carrying forward
established progress.

\begin{figure*}[t]
\centering
\includegraphics[width=0.8\textwidth]{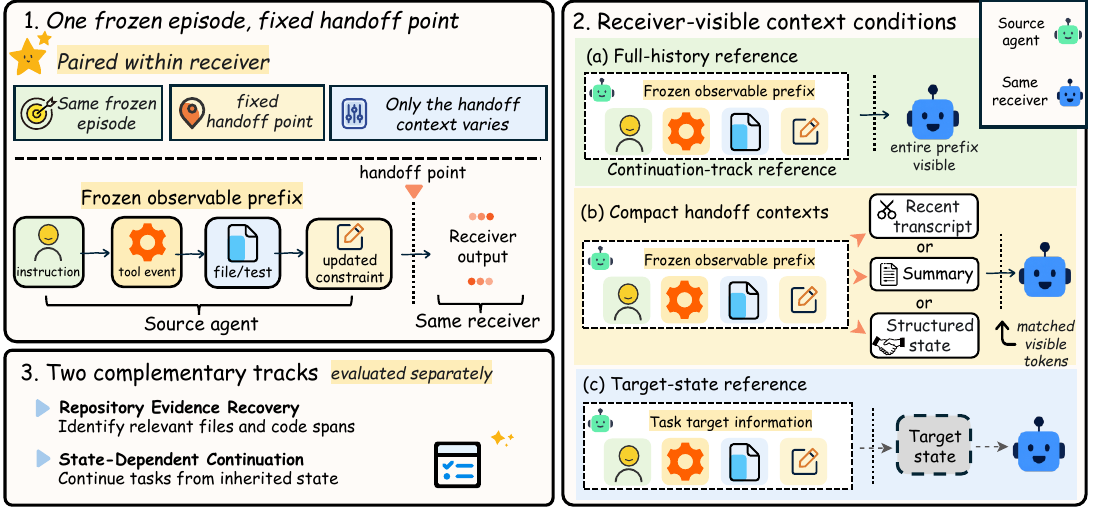}
\caption{Overview of HSC-Bench.
Shared source histories and fixed handoff points support
comparisons across receivers and paired comparisons of
context conditions within each receiver.
Recent transcripts and structured state are evaluated
on both tracks for all six receivers.
GPT-4.1 constructs narrative summaries and structured state
from observable history, with compact contexts matched
in visible token count.
The continuation track additionally evaluates full history
for all six receivers, and narrative summaries and a
target-state reference for GPT-5.4.
The target-state reference supplies privileged task targets.
The two tracks assess repository evidence recovery and
state-dependent continuation separately.}
\Description{The framework diagram links a fixed source execution to
alternative receiver-visible messages and two evaluation tracks. The caption
specifies the scope and additional information access of reference conditions.}
\label{fig:framework}
\end{figure*}

\section{HSC-Bench}
\label{sec:benchmark}

Each task in HSC-Bench begins with a recorded history of work
performed by a source agent. At a fixed handoff point, we prepare
the context provided to receiving models, using GPT-4.1 as the
fixed model for constructing structured state and narrative
summaries. Receivers share the same source history, handoff
point, and prepared context within each condition, allowing
their performance to be compared on the same inherited work.
The benchmark comprises 300 repository evidence-recovery tasks
and 739 state-dependent continuation tasks
(Figure~\ref{fig:framework}).

\subsection{Repository Evidence Recovery Track}

The repository track evaluates whether a receiving model can
identify code relevant to an unresolved issue from the issue
description and inherited context. We pair issues and code-location
annotations from ContextBench~\cite{li2026contextbench} with GPT-5
trajectories released in CodeTraceBench~\cite{li2026codetracer}.
The resulting 300 tasks span 38 repositories and eight programming
languages.
Appendix~\ref{app:data} details source selection and construction.

We place the handoff after ten complete action--observation pairs
and prepare the receiver's context from the preceding history.
The receiver predicts relevant files and line spans, which are
scored against the withheld annotations using line-F1.
The annotations cover the whole issue and may include relevant
locations that do not appear in the pre-handoff history.

To examine how recovery performance varies with the visibility
of these historical cues, we group tasks according to whether
annotated file paths appear in the pre-handoff history and
the recent transcript. In 217 tasks, at least one annotated
path appears in the history but none remains in the recent
transcript; 42 tasks retain at least one annotated path in
the recent transcript; and 41 contain none in the pre-handoff
history.

\subsection{State-Dependent Continuation Track}

The continuation track evaluates how a receiving model carries
forward unfinished work. Its 739 synthetic tasks cover
conversational state, tool use, and reported file-work state.
Following a recorded GPT-5 source history, the receiver continues
through prescribed user and environment messages, with explicit
requirements for its responses and actions.

\textbf{Conversational state.}
Established choices can remain valid even as other facts change.
A project brief, for example, may retain its selected plan
after the user revises the target region. These tasks ask the
receiver to report the resulting conversational state,
preserving prior commitments while incorporating updated
facts and excluding obsolete values.

\textbf{Tool state.}
Completed operations determine which actions remain necessary.
In a stateful order-service simulator, an address update
performed before the handoff should inform how the receiver
handles subsequent requests. The receiver can call tools
and must report the order and operation states. Evaluation
examines both this report and the tool-call record, distinguishing
recognition of a completed update from avoidance of repeating it.

\textbf{File-work state.}
An unfinished coding task leaves a record of progress and
unresolved questions, including file changes, test outcomes,
failed attempts, and working hypotheses. These tasks present
that record through messages and ask the receiver to maintain
a structured progress report as further information arrives.
The report must preserve the required work details and edit
boundaries. The evaluated output is this report; receivers
do not modify files or execute tests.

Each task is evaluated using a fixed set of binary criteria.
Each criterion checks whether a specific requirement concerning
task state, actions, or the requested output is satisfied,
using the relevant response fields or interaction records.
The task score is the proportion of criteria satisfied.
Table~\ref{tab:composition} reports the number of tasks in each
type and the number of criteria applied to each task.
Appendix~\ref{app:criteria} describes the scoring criteria.

\begin{table}[t]
\caption{Controlled tasks and per-task scoring criteria.}
\label{tab:composition}
\centering
\scalebox{0.85}{%
\renewcommand{\arraystretch}{1.15}%
\begin{tabular*}{\linewidth}{@{}l@{\extracolsep{\fill}}rr@{}}
\toprule
\textbf{Task type} & \textbf{Tasks} & \textbf{Criteria} \\
\midrule
Conversational state & 205 & 17 \\
Tool state & 386 & 10 \\
File-work progress & 148 & 8 \\
\midrule
\textbf{Total} & \textbf{739} & --- \\
\bottomrule
\end{tabular*}
}
\end{table}

\subsection{Task Construction and Screening}

For the continuation track, GPT-5.4 generates candidate task
specifications and supplies diagnostic responses and screening
judgments. Each specification contains user and environment
messages, target states, and supporting information.
We check these specifications for internal consistency,
clarity, and plausibility before assessing their dependence
on prior information.

Screening asks whether a task requires the intended history
and whether its targets are attainable with sufficient context.
We probe the former using only the final request, user messages,
or output schema, and the latter using full history or target-state
information. Recent-transcript probes additionally assess
difficulty under limited context. Of 1,130 candidates, 739 meet
the screening criteria and enter the continuation track.
GPT-5 then generates the shared source histories used for
receiver evaluation. Appendix~\ref{app:data} provides the
screening thresholds.

\section{Evaluation Protocol}
\label{sec:protocol}

We compare receiving models on the same recorded tasks and
examine how their performance changes under different
handoff context conditions.

\subsection{Models and Context Conditions}

The shared evaluation compares six receivers: GPT-5.4,
Claude Sonnet 5, DeepSeek V3.2, Qwen 3.5 Plus, Gemini 3.5 Flash Lite, and Kimi K2. All six receive recent transcripts and structured state in both tracks. Recent transcripts retain the end of the pre-handoff history in its original wording; structured state organizes relevant facts, commitments, prior actions, and remaining work. On the continuation track, all six are also evaluated with the complete pre-handoff history.

Supplementary comparisons on GPT-5.4 cover all 739 continuation
tasks and add two conditions: narrative summaries, which
condense prior work into prose, and a target-state reference,
which supplies privileged task targets for diagnostic evaluation. Thus, GPT-5.4 is evaluated with all three compact representations and both reference contexts. We additionally evaluate GPT-5 continuing from its own full history as a source-model reference.

GPT-4.1 constructs summaries and structured state from the task
description and observable history, without access to reference
annotations or target values. For file-work tasks, only
structured-state preparation additionally receives the required
output-field names. Recent transcripts and summaries are matched to the structured-state message's visible token count for each task. Within-model comparisons keep the task request and scoring rules fixed, varying the information selected and how it is presented. Appendix~\ref{app:data} provides model
settings and context preparation details.

\begin{table*}[t!]
\centering
\small
\setlength{\tabcolsep}{3pt}
\caption{Receiver performance on the Repository Evidence Recovery
and State-Dependent Continuation tracks, abbreviated as
Repository and State-Dependent in the table.
Repository scores are line-F1 on 300 tasks; continuation
results report mean scores on a 0--100 scale and
all-criteria pass rates in percent on 739 tasks.
Recent, State, and Full denote recent transcripts,
structured state, and full history, respectively.
The task-type breakdown reports structured-state mean
scores for conversational, tool, and file-work state.
Bold indicates the highest value in each column.}
\label{tab:benchmark-overview}

\begin{tabular}{@{}lrrrrrrrrrrr@{}}
\toprule
& \multicolumn{2}{c}{Repository}
& \multicolumn{9}{c}{State-Dependent} \\
\cmidrule(lr){2-3}\cmidrule(lr){4-12}

& \multicolumn{2}{c}{Line-F1}
& \multicolumn{3}{c}{Mean score}
& \multicolumn{3}{c}{Pass (\%)}
& \multicolumn{3}{c}{\shortstack{State by task type\\(Mean score)}} \\
\cmidrule(lr){4-6}\cmidrule(lr){7-9}\cmidrule(lr){10-12}

Receiver
& Recent & State
& Recent & State & Full
& Recent & State & Full
& Conv. & Tool & File \\
\midrule

GPT-5.4
& 0.075 & 0.229
& \textbf{89.73} & \textbf{92.76} & \textbf{94.51}
& \textbf{44.7} & \textbf{52.1} & \textbf{57.9}
& \textbf{87.63} & \textbf{96.81} & \textbf{89.27} \\

Claude Sonnet 5
& 0.121 & \textbf{0.263}
& 87.19 & 91.05 & 92.77
& 41.1 & 50.2 & 52.1
& 86.74 & 96.61 & 82.52 \\

DeepSeek V3.2
& 0.033 & 0.189
& 85.29 & 89.54 & 92.65
& 35.0 & 47.2 & 55.3
& 85.45 & 95.18 & 80.49 \\

Qwen 3.5 Plus
& 0.089 & 0.250
& 86.71 & 89.84 & 92.37
& 36.7 & 47.2 & 51.6
& 85.80 & 95.98 & 79.39 \\

Gemini 3.5 Flash Lite
& \textbf{0.147} & 0.261
& 77.96 & 85.83 & 90.65
& 30.7 & 42.9 & 53.9
& 86.48 & 90.54 & 72.64 \\

Kimi K2
& 0.031 & 0.192
& 82.78 & 79.45 & 88.90
& 27.5 & 36.5 & 42.2
& 83.87 & 81.81 & 67.15 \\

\bottomrule
\end{tabular}
\end{table*}

\begin{figure*}[t]
    \centering
    \includegraphics[width=0.9\textwidth]{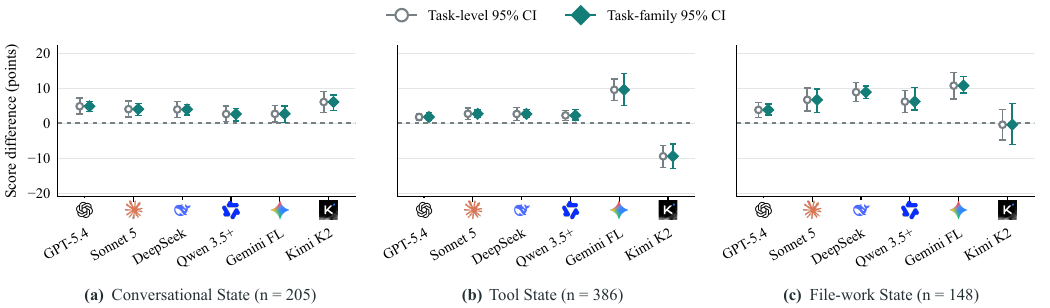}
    \caption{Effects of handoff context within the
    State-Dependent Continuation Track.
    Panels show conversational, tool, and file-work
    state tasks. Points represent paired mean score
    differences, computed as structured state minus
    recent transcript; positive values favor
    structured state. Error bars show 95\% bootstrap
    confidence intervals from resampling tasks or
    task families. Sample sizes appear below each
    panel. Model abbreviations correspond to
    Table~\ref{tab:benchmark-overview}.}
    \label{fig:profiles}
\end{figure*}

\subsection{Evaluation Metrics}

For repository evidence recovery, we report line-F1 as the
primary metric and file-F1 as a complementary measure of
whether receivers identify the relevant files. Both metrics
are computed per task and averaged across tasks.

For state-dependent continuation, each task $i$ has $K_i$
binary criteria. Let $y_i$ denote the receiver's response
or interaction trace, and let $c_{ij}(y_i)$ equal 1 if
criterion $j$ is satisfied and 0 otherwise.
Each criterion uses the designated response fields or
interaction records, together with the fixed source
history and task targets where required.
The task score is the fraction of criteria satisfied:
\begin{equation}
s_i(y_i)=\frac{1}{K_i}\sum_{j=1}^{K_i}c_{ij}(y_i),
\qquad c_{ij}(y_i)\in\{0,1\}.
\label{eq:rubric}
\end{equation}
Appendix~\ref{app:criteria} details the criteria and their
passing conditions.

For a fixed receiver and context condition, we report
the mean task score $\bar{s}$ and the empirical
all-criteria pass rate $\hat{p}$ over $N$ evaluated tasks:
\begin{equation}
\begin{aligned}
\bar{s} &= \frac{1}{N}\sum_{i=1}^{N}s_i(y_i),\\
\hat{p} &= \frac{1}{N}\sum_{i=1}^{N}
\mathbf{1}\!\left[s_i(y_i)=1\right].
\end{aligned}
\label{eq:aggregate_metrics}
\end{equation}
Here $\mathbf{1}[\cdot]$ equals 1 when its condition holds
and 0 otherwise. We label these metrics Mean and Pass
in tables, respectively.
The mean score captures partial satisfaction of task
requirements, whereas the pass rate requires every
criterion within a task to be satisfied.

Overall continuation metrics weight all 739 tasks equally,
rather than weighting the three task types equally.
We report continuation scores and pass rates on a 0--100
scale and repository F1 scores on a 0--1 scale.

\subsection{Statistical Analysis}

To measure how inherited context affects performance,
we compare each receiver's scores on the same tasks
under different context conditions.
Let $y_i^{(S)}$ and $y_i^{(R)}$ denote a fixed receiver's
outputs on task $i$ under structured state and recent
transcripts, respectively.
We use $s_i$ to denote line-F1 for repository tasks and
the criterion-based score in Equation~\ref{eq:rubric}
for continuation tasks.
For $N$ paired tasks, the task-level difference $d_i$
and its mean $\widehat{\Delta}$ are
\begin{equation}
\begin{aligned}
d_i &= s_i(y_i^{(S)})-s_i(y_i^{(R)}),\\
\widehat{\Delta} &= \frac{1}{N}\sum_{i=1}^{N}d_i.
\end{aligned}
\label{eq:paired}
\end{equation}
Positive differences favor structured state.
For GPT-5.4, the comparison with narrative summaries
uses the same within-task pairing.

We report 95\% percentile bootstrap confidence intervals
from 20,000 resamples. Repository analyses resample
whole repositories while retaining equal task weights
within each draw. Continuation analyses resample paired
tasks; an additional analysis resamples whole task
families to assess sensitivity to dependence among
tasks sharing a scenario.
All resampling preserves the pairing between context
conditions.

\section{Results}
\label{sec:results}

\subsection{Model Performance Across the Two Tracks}

As shown in Table~\ref{tab:benchmark-overview},
Claude Sonnet 5 and Gemini achieve the highest repository
evidence-recovery scores with structured state, reaching
line-F1 values of 0.263 and 0.261, respectively.
Qwen follows at 0.250, ahead of GPT-5.4 at 0.229,
while Kimi and DeepSeek obtain similar scores of
0.192 and 0.189.
The leading pair also achieves the highest repository
scores with recent transcripts, although Gemini ranks
ahead of Claude in that condition.

\begin{figure}[t]
\centering
\includegraphics[width=0.8\linewidth]{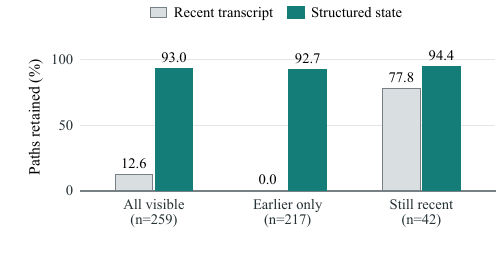}
\caption{Evidence available at takeover. Path retention is the fraction of
previously observed, annotated target paths retained in the received context.
Bars average this fraction over tasks. The 259 eligible tasks
comprise 217 earlier-only and 42 still-recent tasks; retention is undefined
for the remaining 41. Earlier-only transcript retention is zero by definition.}
\Description{Path retention is 12.6 versus 93.0 percent overall, zero versus
92.7 percent for earlier-only evidence, and 77.8 versus 94.4 percent for
still-recent evidence, for recent transcripts and structured state.}
\label{fig:visibility}
\end{figure}

GPT-5.4 leads state-dependent continuation in both mean
score and all-criteria pass rate under all three shared
context conditions. Claude has the second-highest mean
throughout. With structured state, their mean scores
are 92.76 and 91.05, and their pass rates are 52.1\%
and 50.2\%, respectively.
Qwen and DeepSeek form a closely matched pair, with
mean scores of 89.84 and 89.54 and pass rates that both
round to 47.2\%. Gemini and Kimi obtain lower continuation
means, despite Gemini's strong repository recovery.
Thus, the models leading code localization are not
identical to those leading continuation.

The task-type breakdown shows that the gaps between
receivers are uneven across continuation requirements.
Under structured state, GPT-5.4 exceeds Claude by only
0.20 points on tool state, but by 6.75 points on reported
file-work state.
Gemini's conversational-state score is also close to
Claude's, differing by only 0.26 points, while its tool
and file-work scores are lower by 6.07 and 9.88 points,
respectively.
Gemini's lower overall continuation score therefore
coexists with conversational-state performance close
to that of the second-ranked receiver.
The comparison distinguishes models with similar
conversational-state scores but different performance
on operation state and unfinished file-work reports.

\subsection{Effects of Handoff Context}

Structured state improves repository evidence recovery
across all six receivers.
Table~\ref{tab:real-primary} shows line-F1 gains of
0.114--0.162 over recent transcripts, with all
repository-level bootstrap intervals above zero.
On the continuation track, five receivers also improve
in overall mean score. We next examine how these gains
are distributed across task types and individual tasks.

The task types that benefit most differ across receivers.
Figure~\ref{fig:profiles} shows that Gemini gains most
on file-work state, improving by 10.73 points, compared
with 2.67 points on conversational state.
GPT-5.4 instead gains most on conversational state,
with a 4.85-point increase, while its tool-state gain
is 1.76 points.
For Kimi, the difference extends to the direction of
the effect: conversational-state scores rise by
6.05 points, whereas tool-state scores fall by
9.43 points. Both directions persist under task-level
and task-family resampling.
These comparisons show that structured state does not
benefit the same task requirements equally across
receivers. An overall mean can therefore conceal
which parts of continuation improve and which remain
difficult for a particular model.

\begin{figure*}[t]
\centering
\includegraphics[width=0.85\textwidth]{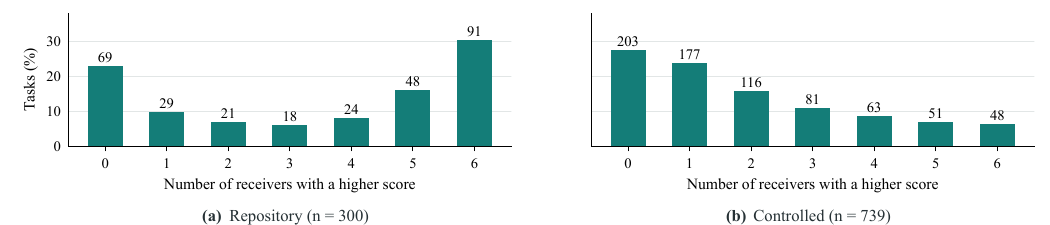}
\caption{Shared and receiver-specific improvements. Each bin counts tasks
on which structured state scores higher than the recent transcript for the
given number of receivers. Heights are task percentages; labels give
repository and controlled task counts under the main scoring rules.}
\Description{Two histograms show the number of benefiting receivers, from
zero to six, on 300 repository tasks and 739 controlled tasks.}
\label{fig:agreement}
\end{figure*}

Agreement in average gains also does not imply agreement
on the tasks that benefit.
Figure~\ref{fig:agreement} shows that 231 of the
300 repository tasks improve for at least one receiver,
but only 91 improve for all six.
For continuation, 536 of the 739 tasks improve for
at least one receiver, while only 48 improve for all six.
Thus, the positive average effects observed across
multiple receivers arise from partially overlapping
sets of tasks.
In continuation, most tasks with an observed gain
benefit only a subset of models.
The usefulness of a handoff context therefore depends
on the receiving model even when the underlying task
and source history are held fixed.

\begin{figure*}[t]
\centering
\includegraphics[width=0.85\textwidth]{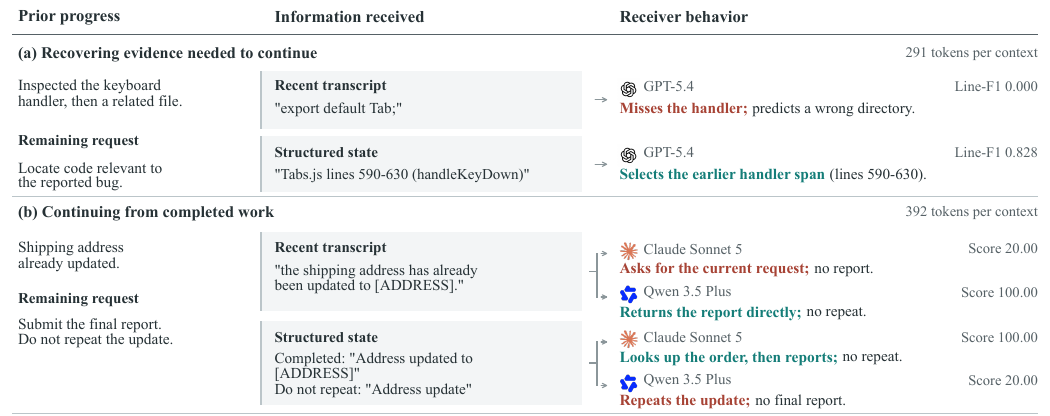}
\caption{Evidence recovery and subsequent action in task takeover.
Read left to right: prior progress and the remaining request, information
received, and actual behavior. \textbf{(a)} GPT-5.4 locates the earlier
keyboard handler when its location is retained. \textbf{(b)} Both contexts
record the completed address update, yet receivers submit the report,
ask for the request again, or repeat the update. Input excerpts are quoted;
behavior descriptions are condensed from recorded outputs. Gray values
give complete-prediction line-F1 in (a) and mean scores (0--100) over ten
requirements in (b). Appendix~\ref{app:cases} details case selection and
action sequences.}
\Description{Two cases read from prior progress through received information
to receiver behavior. GPT-5.4 misses an earlier code location under recent
transcripts and recovers it under structured state. After a completed address
update, Claude asks for the request under recent transcripts and reports under
structured state; Qwen reports under recent transcripts and repeats the update
under structured state.}
\label{fig:cases}
\end{figure*}

\begin{table}[t]
\centering
\small
\setlength{\tabcolsep}{5pt}
\caption{GPT-5.4 across context conditions on all 739 controlled tasks. Mean scores use a 0--100 scale; completion requires all task criteria over the full record. Target state supplies privileged target information.}
\label{tab:references}
\scalebox{0.9}{%
\begin{tabular}{@{}lrr@{}}
\toprule
Context condition & Mean score & Completion (\%) \\
\midrule
Recent transcript & 89.73 & 44.7 \\
Narrative summary & 87.34 & 42.8 \\
Structured state & 92.76 & 52.1 \\
\midrule
Full history & 94.51 & 57.9 \\
Target state & 98.30 & 85.3 \\
\bottomrule
\end{tabular}
}
\end{table}

\subsection{Evidence Recovery Beyond Recent History}

Structured state provides the largest recovery gains
when relevant evidence has left the recent context.
On the 217 tasks whose annotated target paths appear
earlier in the source history but not in the recent
transcript, line-F1 gains range from 0.132 to 0.189.
Table~\ref{tab:visibility-estimates} shows that this
group has the largest estimated gain for every receiver,
with all repository-level bootstrap intervals above zero.
Figure~\ref{fig:visibility} shows the corresponding
difference in evidence retention: across the 259 tasks
with target paths observed in the source history,
structured state retains an average of 93.0\% of those
paths, compared with 12.6\% for recent transcripts.
The recovery advantage is therefore strongest where
retaining only the latest interactions leaves out
previously encountered evidence.

Figure~\ref{fig:cases}(a) illustrates how an earlier
location contributes to the receiver's prediction.
The source model inspects a keyboard handler before
moving to a related file. With the recent transcript,
GPT-5.4 misses the handler; with structured state,
which retains its location, GPT-5.4 includes it in
the predicted evidence. The other five receivers also
obtain higher line-F1 on this task.
The example shows why the latest focus of an
investigation may be insufficient for takeover:
a location encountered earlier can remain relevant
to the unresolved request.

Retaining a relevant file, however, does not by itself
resolve which lines should be returned.
To examine this distinction, we apply a fixed rule
that directly extracts explicit file paths and line
ranges from structured state, as detailed in
Appendix~\ref{app:literal-readout}.
This rule achieves file-F1 of 0.578, within the
receivers' range of 0.531--0.599, but its line-F1
is only 0.080, below all six receivers at 0.189--0.263.
It produces valid line predictions on 90 of the
300 tasks.
Thus, the explicit locations in structured state
already support substantial file-level recovery,
while the receivers achieve finer-grained localization
than this direct extraction rule.

\subsection{Continuing from Recorded Progress}
\label{sec:completion}

A record of completed work should change what the
receiver does next. Our continuation traces show
that this does not always happen, even when the
completion record is explicitly retained.
Figure~\ref{fig:cases}(b) illustrates an order task
in which the address has already been updated
and only the final report remains.
Both context conditions preserve this information,
yet the receivers take different courses of action.
With structured state, Claude checks the order
and submits the report without repeating the update.
Qwen instead repeats the completed update and ends
without a final report, although it submits the report
when given recent transcripts.
The failure occurs after the necessary completion
information has reached the receiver.

To examine whether receivers repeat an operation
whose completion is visible, we inspect 19 tasks
where both contexts explicitly retain the completed
update and its address.
Across six receivers, two of the 114 structured-state
continuations repeat the update, compared with none
under recent transcripts.
Both repetitions occur on the illustrated task,
under DeepSeek and Qwen.
Appendix~\ref{app:receiver-actions} reports action
counts across the broader task set.
These observations identify a specific failure:
an explicit record that an operation is complete
does not always prevent the receiver from executing
it again.

Criterion-level results distinguish reporting completed work
from satisfying the action constraints.
Across receivers, completed-action reporting improves from
77.3\% to 87.3\%, while satisfaction of the update-count
criterion changes from 95.5\% to 95.3\%.
This criterion counts calls across both source and receiver
histories; Appendix~\ref{app:criterion-analysis} provides
the full breakdown.

Together, the two cases illustrate complementary requirements
for task takeover. Retaining an earlier code location can
support evidence recovery, while retaining a completion record
does not always prevent repetition of the completed operation.
The two tracks examine both recovery of relevant information
and continuation consistent with inherited progress.

\subsection{Context Representation and Access to Task State}

The supplementary comparisons on GPT-5.4 examine
whether the benefits of structured state extend
beyond comparison with recent transcripts.
Table~\ref{tab:references} shows that structured
state outperforms narrative summaries by 5.42 points
in mean score and 9.3 percentage points in
all-criteria pass rate.
Both representations condense prior interactions,
but the narrative summary does not match the
continuation performance of structured state.
The difference motivates a closer examination
of their length and source information.

Some narrative summaries are truncated to meet
the token budget, and structured-state preparation
additionally receives output-field names for
file-work tasks.
We therefore examine the 429 conversational and
tool tasks with untruncated summaries and equivalent
source access.
Structured state retains a mean advantage of
3.04 points, with a task-level 95\% bootstrap
confidence interval of [2.07, 4.03].
The advantage also remains positive under
task-family resampling, as reported in
Appendix~\ref{app:summary-budgets}.
Thus, summary truncation and the additional
file-work information do not fully account for
the observed advantage.
This comparison concerns the information selected
and its presentation together.

The reference contexts further distinguish access
to the interaction record from access to the
expected task state.
With full history, GPT-5.4 achieves an all-criteria
pass rate of 57.9\%; supplying privileged target-state
information raises it to 85.3\%.
Full history preserves the interactions from which
the receiver must determine the current state
and remaining requirements, whereas the target-state
reference supplies the expected targets directly.
The gap shows that removing history truncation
alone does not eliminate continuation failures.
Together with the action analysis, these results
highlight the importance of how the receiver uses
the supplied information to determine what to do next.

\begin{table}[t]
\centering
\small
\setlength{\tabcolsep}{3pt}
\caption{Completion differences on 739 tasks (structured state minus recent transcript, percentage points). Benchmark uses the criteria in Section~\ref{sec:protocol}; + status match adds order-status matching in the tool-observation field. Both columns give task-family 95\% intervals.}
\label{tab:completion-diagnosis}
\scalebox{0.85}{%
\begin{tabular}{@{}lrr@{}}
\toprule
Receiver & Benchmark & + status match \\
\midrule
GPT-5.4 & +7.44 [3.11, 12.59] & +6.09 [1.79, 11.19] \\
Claude & +9.07 [5.26, 13.32] & -9.88 [-18.49, 0.43] \\
DeepSeek & +12.18 [8.31, 15.60] & -7.58 [-14.50, 0.81] \\
Qwen & +10.55 [5.26, 15.41] & -6.09 [-10.98, -0.15] \\
Gemini & +12.18 [5.71, 17.96] & -3.92 [-7.92, 0.42] \\
Kimi & +9.07 [5.14, 12.76] & -5.28 [-10.28, 0.74] \\
\bottomrule
\end{tabular}
}
\end{table}

\subsection{When Mean Scores and Pass Rates Diverge}
\label{sec:criterion-satisfaction}

The effect of handoff context depends on whether
we measure average criterion satisfaction or
the proportion of tasks satisfying every criterion.
Table~\ref{tab:completion-diagnosis} shows that
structured state improves the all-criteria pass
rate for all six receivers, with gains of
7.44--12.18 percentage points.
All task-family bootstrap intervals for these
gains are above zero.
For five receivers, higher pass rates accompany
higher mean scores. Kimi presents a different
pattern: structured state yields 67 more tasks
that pass every criterion, despite a 3.33-point
decrease in its mean score.
Its lower mean therefore does not imply that
fewer tasks meet all evaluation requirements.

The distinction also matters when interpreting
absolute performance.
With structured state, GPT-5.4 achieves a mean
score of 92.76, but only 52.1\% of tasks pass
every criterion. Claude similarly obtains a mean
of 91.05 and a pass rate of 50.2\%.
These models satisfy most criteria on average,
while approximately half of their tasks still
have at least one unmet requirement.

\section{Discussion}

A handoff message can correctly record that an
operation is complete while the receiving model
still performs it again.
This matters because a completion record changes
what the model needs to do next: it can move on
to the unfinished work.
Checking whether the record survives the handoff
therefore answers only part of the evaluation question.
Inspecting the next action helps locate the failure:
was the completion record missing, or did the receiver
repeat the operation despite having that record?
This distinction identifies whether an observed
problem concerns the information passed to the
receiver or its use during continuation.

The receiver comparisons also have implications
for choosing how to pass work between models.
When the remaining task is to locate relevant code,
the useful information may be a file path or a
previously inspected function.
When the remaining task is to report an operation's
outcome, the receiver needs to use the completion
record to decide what to report and whether another
action is necessary.
These requirements give different purposes to
the information retained at handoff.
Comparing representations with the intended receiver
and remaining task therefore provides a more useful
basis for choosing a handoff format than evaluating
the message in isolation.

Repeated handoffs introduce a further problem:
one receiver's output becomes part of the next
receiver's input.
For example, if a model incorrectly reports that
an operation is complete, a later model could skip
work that still needs to be done.
The single-handoff setting studied here provides
a starting point for examining such errors.
An extension could follow the same task through
several receivers and compare each completion
record with the actions actually executed.
This would reveal where an incorrect record first
appears, whether later models act on it, and
whether they correct it before passing the task onward.

\section{Conclusion}

We introduced HSC-Bench to evaluate how models take
over unfinished work through two complementary
tracks: recovering relevant repository evidence
and continuing from inherited task state.
The benchmark connects the context available at
handoff with checks of the receiver's predictions,
responses, and actions.
Our evaluation shows that structured state improves
evidence recovery across all six receivers, with
the largest gains on tasks whose relevant paths
have disappeared from recent transcripts.
However, the models that lead evidence recovery
do not necessarily lead task continuation.
Action records also reveal cases in which receivers
repeat completed operations despite receiving
explicit completion records.
These findings distinguish recovering prior
information from using it to continue existing work.
Assessing task takeover requires checking both
what the receiver can recover and whether its
next actions respect what has already been done.

\clearpage
\setlength{\bibsep}{3.5pt}
\bibliographystyle{ACM-Reference-Format}
\bibliography{references}

\appendix
\raggedbottom
\setcounter{dbltopnumber}{2}
\renewcommand{\dbltopfraction}{0.95}
\renewcommand{\textfraction}{0.04}
\captionsetup{skip=5pt}
\setlength{\floatsep}{4pt plus 1pt minus 1pt}
\setlength{\textfloatsep}{6pt plus 1pt minus 1pt}
\setlength{\intextsep}{4pt plus 1pt minus 1pt}
\setlength{\dbltextfloatsep}{6pt plus 1pt minus 1pt}
\setlength{\dblfloatsep}{6pt plus 1pt minus 1pt}

\section{Data and Evaluation Details}
\label{app:data}

\textbf{Repository selection.}
We align CodeTraceBench trajectories~\cite{li2026codetracer}
with ContextBench annotations~\cite{li2026contextbench}.
A fixed list of 500 tasks yields 445 eligible trajectories for
327 tasks. Manual inspection and deterministic ordering retain
300, excluding calibration, evidence-leakage, and repository-mismatch
cases. Their GPT-5 histories comprise 193 OpenHands
trajectories~\cite{wang2025openhands}, 52 SWE-agent
trajectories~\cite{yang2024sweagent}, and 55 mini-SWE-agent
trajectories.

\textbf{Repository scoring.}
Predictions may contain at most 12 files and 12 line spans
per file. Line numbers must be positive, and span endpoints
are inclusive. Overlapping spans are merged, and unsupported
predictions remain in the precision denominator.
Invalid predictions are scored as empty.
Gain, tie, and loss counts use unrounded scores with
a numerical tolerance of $10^{-12}$.

\textbf{Continuation selection.}
Screening retains 205 of 250 conversational candidates, 386 of
560 tool candidates, and 148 of 320 file-work candidates.
These span five, seven, and five scenario families, respectively.
On 1--5 scales, plausibility must reach 4.0 for hard tasks or
3.5 otherwise, and ambiguity must not exceed 2.5.
Specifications must contain consistent targets and supporting
information. For easy tasks, probe-score ceilings for the final
request, user messages, output schema, and recent transcript are
0.60, 0.80, 0.50, and 0.95; for other tasks, they are 0.35, 0.55,
0.25, and 0.85. Target-state and full-history floors are 0.70 and
0.60 for file-work tasks, or 0.85 and 0.80 otherwise.
Recent-transcript scores must exceed 0.05.
The probes use 900-character contexts and specification-based
checks to select tasks; the criteria in
Appendix~\ref{app:criteria} evaluate the subsequent source
and receiver histories.

\textbf{Context preparation.}
GPT-4.1 prepares structured state with character targets of 1,600
for conversational and tool tasks and 1,800 for file-work and
repository tasks. Repository state must be valid JSON, contain
at most 4,000 characters, and compress its source at least fivefold.
Narrative summaries preserve the same kinds of observed facts in
prose. Their character cap equals the structured message's length,
bounded to 200--1,600 characters. Recent transcripts are truncated
to the visible token budget; summaries receive whitespace padding
after character truncation. Counts use the \texttt{o200k\_base}
tokenizer.

\textbf{Model settings.}
Receiver versions are \nolinkurl{gpt-5.4},
\nolinkurl{claude-sonnet-5},
\nolinkurl{deepseek-v3.2},
\nolinkurl{qwen3.5-plus-2026-02-15},
\nolinkurl{gemini-3.5-flash-lite},
and \nolinkurl{kimi-k2-250905}.
The continuation source is \nolinkurl{gpt-5-2025-08-07}.
Temperature is 0 and top-$p$ is 1, with output limits of 1,536
tokens for continuation and 1,024 for repository tasks.
GPT-5 and Gemini use minimal reasoning; Qwen disables thinking.
Tool continuations allow four model calls, with operations for
order lookup, policy inspection, address updates, and credit calculation.

\begin{table*}[t]
\centering
\small
\setlength{\tabcolsep}{2pt}
\renewcommand{\arraystretch}{1.0}
\caption{Criterion satisfaction rates (\%), averaged equally
across six receivers. Recent and State denote recent transcripts
and structured state. Each row contributes one binary score;
passing conditions are defined in Appendix~\ref{app:criteria}.}
\label{tab:criteria}

\begin{minipage}[t]{0.32\textwidth}
\begin{tabular*}{\linewidth}
{@{}ll@{\extracolsep{\fill}}rr@{}}
\toprule
\multicolumn{4}{@{}l}{\textbf{Conversational state} (205)} \\
\midrule
 & Requirement & Recent & State \\
\midrule
C1  & Primary choice        & 83.6 & 93.1 \\
C2  & Backup choice         & 85.3 & 94.8 \\
C3  & Status label          & 78.2 & 81.1 \\
C4  & Review focus          & 80.1 & 88.8 \\
C5  & Audience              & 79.5 & 86.8 \\
C6  & Continuation strategy & 85.4 & 94.1 \\
C7  & Fallback boundary     & 85.2 & 91.5 \\
C8  & Quality standard      & 88.6 & 94.4 \\
C9  & Current color         & 93.8 & 97.1 \\
C10 & Current region        & 95.9 & 97.3 \\
C11 & Locale                & 83.8 & 89.3 \\
C12 & Accumulated total     & 75.0 & 68.4 \\
C13 & Language field        & 62.2 & 51.5 \\
C14 & Style field           & 65.3 & 84.6 \\
C15 & Risk field            & 55.6 & 52.8 \\
C16 & Output fields         & 99.0 & 99.0 \\
C17 & Obsolete values       & 97.1 & 97.3 \\
\bottomrule
\end{tabular*}
\end{minipage}
\hfill
\begin{minipage}[t]{0.32\textwidth}
\begin{tabular*}{\linewidth}
{@{}ll@{\extracolsep{\fill}}rr@{}}
\toprule
\multicolumn{4}{@{}l}{\textbf{Tool state} (386)} \\
\midrule
 & Requirement & Recent & State \\
\midrule
T1  & Order identifier       & 98.8 & 95.7 \\
T2  & Order status           & 93.6 & 93.3 \\
T3  & Tracking identifier    & 93.9 & 92.3 \\
T4  & Customer tier          & 95.0 & 93.7 \\
T5  & Refund eligibility     & 94.6 & 93.7 \\
T6  & Update completion      & 82.8 & 90.6 \\
T7  & Completed actions      & 77.3 & 87.3 \\
T8  & Non-repeatable actions & 81.0 & 86.4 \\
T9  & Update-call count      & 95.5 & 95.3 \\
T10 & Update permission      & 100.0 & 100.0 \\
\bottomrule
\end{tabular*}
\end{minipage}
\hfill
\begin{minipage}[t]{0.32\textwidth}
\begin{tabular*}{\linewidth}
{@{}ll@{\extracolsep{\fill}}rr@{}}
\toprule
\multicolumn{4}{@{}l}{\textbf{File-work state} (148)} \\
\midrule
 & Requirement & Recent & State \\
\midrule
F1 & Relevant files          & 95.9 & 98.0 \\
F2 & Test observations       & 12.6 & 38.9 \\
F3 & Rejected attempts       & 68.0 & 77.8 \\
F4 & Working hypothesis      & 75.6 & 78.8 \\
F5 & Reported test success   & 68.0 & 66.2 \\
F6 & Rejected plan exclusion & 68.0 & 77.8 \\
F7 & Edit boundaries         & 94.8 & 98.0 \\
F8 & Output fields           & 97.9 & 93.1 \\
\bottomrule
\end{tabular*}
\end{minipage}
\end{table*}

\begin{table*}[t]
\centering
\small
\setlength{\tabcolsep}{3pt}
\caption{Repository recovery overall and by visibility of
annotated paths. File-F1 uses structured state; differences
are structured state minus recent transcript, with
repository-level 95\% bootstrap intervals.
The three visibility groups partition all 300 tasks.}
\label{tab:real-primary}
\label{tab:visibility-estimates}
\label{tab:literal-readout}

\begin{tabular*}{\textwidth}
{@{}l@{\extracolsep{\fill}}rcccc@{}}
\toprule
 & & \multicolumn{4}{c}{Line-F1 difference [95\% CI]} \\
\cmidrule(l){3-6}
Receiver & File-F1 & All tasks
& Earlier only (217) & Still recent (42) & Not observed (41) \\
\midrule
GPT-5.4 & 0.5858
& +0.154 [0.107, 0.186]
& +0.182 [0.133, 0.216]
& +0.108 [0.016, 0.195]
& +0.049 [-0.011, 0.101] \\
Claude & 0.5911
& +0.142 [0.095, 0.172]
& +0.168 [0.118, 0.200]
& +0.094 [-0.008, 0.198]
& +0.052 [-0.050, 0.135] \\
DeepSeek & 0.5485
& +0.156 [0.112, 0.186]
& +0.178 [0.127, 0.212]
& +0.147 [0.097, 0.198]
& +0.048 [-0.001, 0.090] \\
Qwen & 0.5988
& +0.160 [0.102, 0.201]
& +0.189 [0.127, 0.233]
& +0.089 [-0.015, 0.172]
& +0.081 [0.010, 0.151] \\
Gemini & 0.5724
& +0.114 [0.079, 0.145]
& +0.132 [0.103, 0.166]
& +0.071 [-0.070, 0.182]
& +0.064 [-0.017, 0.135] \\
Kimi & 0.5308
& +0.162 [0.108, 0.199]
& +0.187 [0.129, 0.224]
& +0.128 [0.020, 0.205]
& +0.063 [0.006, 0.124] \\
\bottomrule
\end{tabular*}
\end{table*}

\begin{figure*}[t]
\centering
\includegraphics[width=\textwidth]
{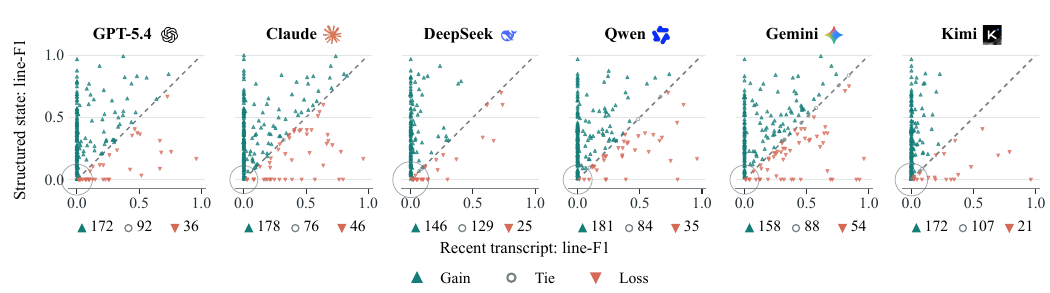}
\caption{Evidence recovery for task takeover
(300 tasks per receiver).
Counts below panels give gains, ties, and losses in line-F1
with structured state relative to recent transcripts.
Point area counts tasks sharing a score pair.}
\label{fig:paired-recovery}
\end{figure*}

\begin{table*}[t]
\centering
\small
\setlength{\tabcolsep}{4pt}
\renewcommand{\arraystretch}{1.05}
\caption{Three recorded continuation examples, abridged from
task histories and GPT-5.4 responses under structured state.
The last column summarizes the observed continuation and
the number of satisfied criteria.}
\label{tab:task-examples}

\begin{tabularx}{\textwidth}
{@{}
>{\raggedright\arraybackslash}p{0.11\textwidth}
>{\raggedright\arraybackslash}X
>{\raggedright\arraybackslash}p{0.14\textwidth}
>{\raggedright\arraybackslash}X
@{}}
\toprule
Task
& State before continuation
& Remaining request
& Observed continuation \\
\midrule
Conversational state
&
Eight choices established, including launch readiness
and no invented dates.
Later updates: teal/APAC $\rightarrow$ orange/EMEA;
total $9\rightarrow14$.
&
Report the current brief state.
&
Retains all eight choices; reports orange, EMEA, and 14;
omits obsolete values.
\textbf{17/17 criteria.} \\
\midrule
Tool state
&
Address update already succeeded for order BX-591204.
Order facts and a do-not-repeat instruction remain relevant.
&
Report the final order state.
&
Reports order facts and the completed update;
marks it as non-repeatable; makes no further update call.
\textbf{10/10 criteria.} \\
\midrule
File-work state
&
\texttt{slugify}: two tests pass, one fails.
Hyphen-replacement loop rejected;
separator normalization proposed.
No patch or successful rerun; test file protected.
&
Report current progress.
&
Retains test observations, rejected attempt, hypothesis,
and edit boundary; does not claim verified success.
\textbf{8/8 criteria.} \\
\bottomrule
\end{tabularx}
\end{table*}

\section{Continuation Criteria and Analysis}
\label{app:criteria}
\label{app:scoring}

Table~\ref{tab:criteria} lists all 17 conversational,
10 tool, and eight file-work criteria.
Reported state fields are extracted from the final response
after removing reasoning blocks.
When a response cannot be parsed, its fields are treated
as empty; action and absence criteria are evaluated separately.
Unless stated otherwise, reported fields must equal their
targets after normalizing case and repeated whitespace.
Each field-based criterion examines its designated field.
The format criteria C16 and F8 require exactly the requested
16 and eight fields, respectively, regardless of field order.

\textbf{Conversational targets.}
For C1--C8, the reference value is the earliest matching
candidate phrase in the chronological assistant messages.
The search includes both source and receiver messages.
A match in the source history therefore determines the
reference; if none occurs there, the first match in the
receiver's continuation supplies it.
A criterion receives no credit when no candidate is matched.
Current facts come from the task specification;
C12 compares the accumulated total as an integer.
Region and locale require nonempty fields and allow either
normalized string to contain the other.
Style accepts exact matching or at least
$\max(1,\lfloor k/2\rfloor)$ target-keyword matches,
where $k$ counts words of at least four characters excluding
``with'' and ``style.''
Low-risk equivalents are R1, low, low risk, routine,
and routine risk.
C17 requires obsolete values to be absent under whole-word
matching, except when a value is contained within a different
active target.
Language, style, and risk criteria assess reported fields,
not the linguistic quality of the surrounding prose.

\textbf{Tool reports and actions.}
Criteria T1--T5 compare the reported order facts with their
target values.
T6 is satisfied when the update-status field explicitly
confirms completion; planned, negated, uncertain, or
contradictory completion claims do not satisfy it.
T7 requires the completed-action field to include the address
update, and T8 requires the non-repeatable-action field to
identify that update.
Both additionally require a successful update in the recorded
history.
T9 is satisfied when the combined source and receiver histories
contain at most one address-update call, including failed calls.
This differs from the receiver-only repetitions examined in
Appendix~\ref{app:receiver-actions}.
T10 requires no address-update calls when the task prohibits
updates; otherwise, it is satisfied.
All evaluated tool tasks satisfy this criterion.

\textbf{File-work reports.}
F1 requires the designated file field to name all required paths.
Test observations, rejected attempts, and hypotheses must each
match in their own field.
A target statement matches either verbatim after normalization
or through at least $\min(3,\max(1,\lfloor k/3\rfloor))$ keyword
matches among its $k$ keywords;
keywords have at least five characters and exclude ``expected,''
``failed,'' ``attempt,'' ``current,'' and ``state.''
F5 evaluates the reported test-success value.
The values true, pass, passed, and 1 are interpreted as positive
and require supporting file, test, or hypothesis content.
The binary target is 1 for the 64 tasks with confirmed passing
tests and 0 for the 61 failing, 11 unverified, and 12 ambiguous
cases. The detailed test-report criteria additionally assess
the recorded observations.
F6 requires the rejected-attempt criterion to be satisfied
and rejects a verbatim rejected plan in the patch summary unless
the summary contains ``do not'' or ``reject.''
F7 also requires F1 and permits protected paths only with a local
statement that they remain unchanged or must not be edited.
These checks assess reported progress and scope, not executed edits.

\newpage
\subsection{Which Requirements Improve?}
\label{app:criterion-analysis}

The criterion-level analysis identifies which requirements
account for the aggregate gains.
In tool tasks, structured state improves reporting of completed
actions by 10.0 percentage points and update status by 7.8 points.
Satisfaction of the update-count criterion changes from
95.5\% to 95.3\%.
Better completion reports therefore coexist with essentially
unchanged compliance on this action-count measure.

For file-work tasks, the largest increase concerns test observations:
satisfaction rises from 12.6\% to 38.9\%, while satisfaction of
the reported test-success criterion changes from 68.0\% to 66.2\%.
Recovering detailed observations and correctly reporting
whether tests have passed are distinct requirements.
Even with structured state, detailed test reports remain
incomplete on many tasks.
Conversational gains also differ by requirement:
style-field matching improves by 19.3 points,
whereas language-field matching falls by 10.7 points.

\section{Supporting Comparisons}
\label{app:additional-results}

\textbf{Repository localization.}
Table~\ref{tab:real-primary} reports file-F1 and paired line-F1
differences overall and by evidence visibility.

Figure~\ref{fig:paired-recovery} shows the paired localization scores.
Across receivers, the number of improving tasks ranges
from 146 to 181 out of 300, while the number of declining
tasks ranges from 21 to 54.
Of the improving tasks, 90--131 move from zero to positive
overlap with the reference code lines.

\textbf{Direct extraction of code locations.}
\label{app:literal-readout}
The comparison in Section~\ref{sec:results} directly extracts full
file paths and positive line spans attached to the same path from
the code-location field of structured state.
It excludes aliases, symbol-only references, and command ranges.
Of 300 messages, 299 yield file predictions and 91 contain
line spans; one exceeds the span limit, leaving 90 valid
line predictions.
File-F1 is 0.5782 and line-F1 is 0.0799.

\subsection{Supplementary Continuation Comparisons}
\label{app:summary-budgets}

\textbf{Summary length and source access.}
The character cap truncates 284 of 739 summaries, leaving 455
intact. Table~\ref{tab:summary-budgets} isolates intact conversational
and tool summaries, whose preparation has the same source access
as structured state.
Padding is the fraction of visible tokens removed when trailing
whitespace is stripped; both counts include the summary label.
The advantage remains positive when this fraction is at most 20\%.

\textbf{Source-model reference.}
GPT-5 continuing from its own full history achieves
a mean score of 93.63 and a pass rate of 53.0\%.

\begin{table}[t]
\centering
\small
\setlength{\tabcolsep}{2pt}
\caption{GPT-5.4 mean-score differences (State minus Summary),
with task-level and task-family 95\% bootstrap intervals.
Intact summaries are untruncated; non-file tasks equalize
source access.}
\label{tab:summary-budgets}

\begin{tabular*}{\linewidth}
{@{}l@{\extracolsep{\fill}}rcc@{}}
\toprule
Subset & Tasks
& \shortstack{Difference\\{[task CI]}}
& Family CI \\
\midrule
All & 739
& +5.42 [4.44, 6.41]
& [2.46, 9.20] \\
Non-file & 591
& +5.84 [4.74, 6.94]
& [2.20, 10.68] \\
Intact non-file & 429
& +3.04 [2.07, 4.03]
& [1.35, 6.39] \\
Padding $\leq20\%$ & 97
& +4.12 [2.02, 6.41]
& [1.22, 8.89] \\
\bottomrule
\end{tabular*}
\end{table}

\section{Task Examples and Action Evidence}
\label{app:cases}
\label{app:additional}
\label{app:examples}

Table~\ref{tab:task-examples} presents three recorded tasks,
showing how receivers retain earlier choices, recognize completed
operations, and report unfinished file work.
The excerpts are abridged from histories and evaluated responses.

Figure~\ref{fig:cases}(a) illustrates a repository location
missed under recent transcripts but recovered under
structured state.
GPT-5.4 predicts 44 lines, including 41 of 55 reference
lines, giving line-F1 of $82/99$.

\newpage
\subsection{Actions After a Completed Update}
\label{app:receiver-actions}

Figure~\ref{fig:cases}(b) illustrates different continuations
after a successful address update.
Qwen repeats the update on its third call, before the
four-call limit, and submits no final report.

Table~\ref{tab:receiver-actions} reports continuations with
update calls, including failed attempts, and successful repeats
for the same order and address.
Source-update groups are disjoint; the 19-task explicit-context
subset overlaps them.
Both contexts identify the completed update and address.
Its two repeats occur under DeepSeek and Qwen on the illustrated
task. Clarification requests count as final responses,
even when they do not provide the requested report.

\par\medskip\noindent
\begin{minipage}{\linewidth}
\captionsetup{type=table}
\centering
\small
\setlength{\tabcolsep}{2pt}
\caption{Action outcomes (State / Recent) across six receivers.
$n$ denotes continuations per condition; outcomes may overlap.}
\label{tab:receiver-actions}
\label{tab:action-strata}

\begin{tabular*}{\linewidth}
{@{}l@{\extracolsep{\fill}}rrrr@{}}
\toprule
Source history & $n$
& \shortstack{Update\\call}
& \shortstack{Same-target\\repeat}
& \shortstack{No final\\response} \\
\midrule
All tool tasks & 2,316 & 39 / 18 & 18 / 12 & 17 / 10 \\
One update & 2,172 & 18 / 14 & 18 / 11 & 14 / 6 \\
No update & 54 & 21 / 3 & 0 / 0 & 3 / 4 \\
Multiple updates & 90 & 0 / 1 & 0 / 1 & 0 / 0 \\
\midrule
Explicit in both & 114 & 2 / 0 & 2 / 0 & 2 / 0 \\
\bottomrule
\end{tabular*}
\end{minipage}

\end{document}